\documentclass[10pt,twocolumn,letterpaper]{article}

\usepackage{cvpr}
\usepackage{times}
\usepackage{epsfig}
\usepackage{graphicx}
\usepackage{amsmath}
\usepackage{amssymb}

\usepackage{amsmath,amsfonts,bm}

\def\eqref#1{equation~\ref{#1}}

\def\1{\bm{1}}

\def\va{{\bm{a}}}
\def\vb{{\bm{b}}}

\def\vg{{\bm{g}}}

\def\vm{{\bm{m}}}

\def\vs{{\bm{s}}}

\def\vv{{\bm{v}}}
\def\vw{{\bm{w}}}
\def\vx{{\bm{x}}}
\def\vy{{\bm{y}}}

\def\mA{{\bm{A}}}

\def\mH{{\bm{H}}}
\def\mI{{\bm{I}}}

\def\mM{{\bm{M}}}

\def\mV{{\bm{V}}}

\DeclareMathAlphabet{\mathsfit}{\encodingdefault}{\sfdefault}{m}{sl}
\SetMathAlphabet{\mathsfit}{bold}{\encodingdefault}{\sfdefault}{bx}{n}

\newcommand{\E}{\mathbb{E}}

\newcommand{\Var}{\mathrm{Var}}

\usepackage{algorithm}
\usepackage{algorithmic}
\usepackage{amsfonts}
\usepackage{mathrsfs}
\usepackage{caption}
\usepackage{subfig}
\usepackage{color}
\usepackage{url}

\usepackage[pagebackref=true,breaklinks=true,letterpaper=true,colorlinks,bookmarks=false]{hyperref}

\cvprfinalcopy 

\def\cvprPaperID{5637} 

\ifcvprfinal\fi
\begin{document}

\title{An Adaptive Remote Stochastic Gradient Method for Training Neural Networks}

\author{Yushu Chen \and Hao Jing \and Wenlai Zhao \and Zhiqiang Liu \and Ouyi Li \and Liang Qiao \and Wei Xue \and Guangwen Yang\thanks{ Corresponding author.}\\
Department of Computer Science and Technology,  \thanks{ Also at National Supercomputing Center in Wuxi, Wuxi, Jiangsu, China.}\\
Tsinghua University, Beijing, China\\
{\tt\small ygw@mail.tsinghua.edu.cn, yschen11@126.com}
}

\maketitle

\begin{abstract}
We present the remote stochastic gradient (RSG) method, which computes the gradients at configurable remote observation points, in order to improve the convergence rate and suppress gradient noise at the same time for different curvatures. RSG is further combined with adaptive methods to construct ARSG for acceleration. The method is efficient in computation and memory, and is straightforward to implement. We analyze the convergence properties by modeling the training process as a dynamic system, which provides a guideline to select the configurable observation factor without grid search. ARSG yields $O(1/\sqrt{T})$ convergence rate in non-convex settings, that can be further improved to $O(\log(T)/T)$ in strongly convex settings. Numerical experiments demonstrate that ARSG achieves both faster convergence and better generalization, compared with popular adaptive methods, such as ADAM, NADAM, AMSGRAD, and RANGER for the tested problems. In particular, for training ResNet-50 on ImageNet, ARSG outperforms ADAM in convergence speed and meanwhile it surpasses SGD in generalization.
\end{abstract}

\section{Introduction and related work}

First-order optimization methods e.g. \cite{Robbins1951,Polyak1964,Bottou2010,Sutskever2013,Kingma2015} are competitive workhorses for training neural networks. Compared with second-order methods e.g.  \cite{Nocedal1980,Martens2010,Byrd2016,Osawa2019,Henriques2019,Anil2020}, they are easier to implement since only the gradients are introduced as input. They also require low computation overheads except computing gradients, which is of the same computational complexity as evaluating the functions. The recent improvements to first-order methods include two important categories \cite{Zhang2019}:

One category is accelerated methods. Sutskever et al. \cite{Sutskever2013} show that the momentum is crucial to improve the performance of SGD. Momentum methods such as heavy ball (HB) \cite{Polyak1964}, amplify steps in low-curvature eigen-directions of the Hessian through accumulation, and allow larger step size for convergence along the high-curvature directions compared with vanilla SGD. They \cite{Sutskever2013} rewrite the Nesterov's accelerated gradient (NAG) \cite{Nesterov1983} into a momentum form, that computes the gradient at an observation point ahead of the current point along the last updating direction. They illustrate that NAG suppresses the step along high curvature eigen-directions in order to prevent oscillations, and converges faster than HB. However, all these approaches are approximation of their original forms derived for exact gradients, without full study on gradient noise. Kidambi et al. \cite{kidambi2018} show the insufficiency of HB and NAG in stochastic optimization, especially for small minibatches. They further present ASGD \cite{Jain2018, kidambi2018} and show improvement over SGD in any information-theoretically admissible regime. However, the method requires tuning of 3 hyper-parameters which vary in large ranges and are difficult to estimate, leading to huge costs that limits its application.

Another category is adaptive gradient methods (hereinafter "adaptive methods"
). Being particularly successful among variants of SGD, these methods scale the gradient elementwise by some forms of averaging of the past gradients. ADAGRAD \cite{Duchi2011} is the first popular adaptive method. It is well-suited for sparse gradients, but suffers from rapid decay of step sizes in cases of non-convex loss functions or dense gradients. Subsequent adaptive methods, such as RMSPROP \cite{Tieleman2012}, ADADELTA \cite{Zeiler2012}, ADAM \cite{Kingma2015}, and NADAM \cite{Dozat2016}, mitigate this problem by using the exponential moving averages of squared past gradients to scale the update, but Reddi et al. \cite{Reddi2018} find that these methods may not converge to optimal solutions even in convex settings, and propose AMSGRAD to fix the problem. Besides, it is widely concerned that the adaptive methods generalize worse than SGD in various models \cite{Wilson2017}. Many approaches e.g.  \cite{Keskar2017,luo2019,Liu2020,Tong2019} are presented to improve their generalization, among which RADAM \cite{Liu2020} rectifies the variance of the adaptive learning rate, and SADAM \cite{Tong2019} calibrates the adaptive learning rate. RANGER \cite{Wright2019} accelerates RADAM by combining it with lookahead optimization \cite{Zhang2019}. 

In order to expedite the convergence of first-order methods, in this paper we propose a \textbf{r}emote \textbf{s}tochastic \textbf{g}radient (RSG) method by generalizing NAG, and combine it with adaptive methods to construct \textbf{a}daptive RSG (ARSG) for further acceleration. We summarize our contributions as follows:

$\bullet$ The methods proposed compute the gradients at configurable remote observation points. Instead of approximating NAG for exact gradients, the observation factor which determines the observation position, is adjustable to accelerate convergence in the stochastic settings.

$\bullet$ In order to reduce the cost of grid search in hyper-parameter tuning, a dynamic system model of the training process is proposed. It guides the selection of the observation factor to decrease both the convergence rate and the accumulation of gradient noise for different curvatures.  

$\bullet$ For further acceleration, ARSG scales the update by an elementwise preconditioner, which is modified to improve generalization and save computation.

\section{Method}

In this section, we propose the RSG method by generalizing NAG with configurable observation distance, and equip it with adaptive preconditioners to construct ARSG.

\textbf{Notation}. We denote $f(\vx)$ as the stochastic function to optimize, where $\vx$ is the parameter vector. $f_{t}(\vx)$ is the evaluation of $f(\vx)$ at the $t$-th iteration with a minibatch of $b$ samples. $d$ is the dimension of vectors and matrices, and ${S_{+}^d}$ is the set of all positive definite $d\times{d}$ matrix. The gradient observation is noisy as $\nabla{f_{t}}(\vx)=\nabla f(\vx)+\zeta_t$, where $\zeta_t$ is the gradient noise. We use $[N]$ to denote the set $\{1,\cdots, N\}$, and use $O(\cdot)$,  $o(\cdot)$, $\Omega(\cdot)$, $\omega(\cdot)$ as standard asymptotic notations. For a vector $\va\in{R^d}$ and a matrices $\mM\in{R^d\times{R^d}}$, $\|\va\|$ is $\|\va\|_2$, $\va/\mM$ is $\mM^{-1}\va$, $\mathrm{diag}(\va)$ is a square diagonal matrix with the elements of $\va$ on the main diagonal, $\mathrm{diag}(\mM)$ is the diagonal vector of $\mM$,  $\mM_i$ is the $i$-th row of $\mM$, and $\sqrt{\mM}$ is $\mM^{1/2}$. For any vectors $\va,\vb\in{R^d}$, we use $\sqrt{\va}$, $\va^2$, $\va/\vb$, and $\max(\va,\vb)$ for elementwise operations. $\mathcal{F}\subset{R^d}$ is the feasible set of points, and the projection operation is defined as  $\Pi_{\mathcal{F},\mA}(y)=\arg \min_{\vx\in{\mathcal{F}}}\|\mA^{1/2}(\vx-\vy)\|$ for $\mA\in{S_{+}^d}$ and  $\vy\in{R^d}$. 

In order to accelerate SGD, HB amplifies the step along low curvature directions through accumulating a momentum across iterations instead of using the stochastic gradient directly, as
\begin{equation}
\begin{aligned}
&\vm_t=\beta_{t}\vm_{t-1}+(1-\beta_{t}){\nabla}{f}_t(\vx_t)\\
&\vx_{t+1}=\vx_t-\alpha_{t}\vm_{t},
\end{aligned}
\label{HBUpd}
\end{equation}
where $\alpha_t$, $\beta_{t}$ are configurable coefficients, $\vm_t$ is the momentum, and $\vm_0=0$. 

Since the updating directions are partially maintained, gradients computed at observation points, which lie ahead of the current point along the last updating direction, contain the predictive information of the forthcoming update.To make full 
use of the predictive information, we compute the gradient at a configurable observation point, and substitute the stochastic gradient with the gradient observation in HB, obtaining the original form of remote stochastic gradient (RSG) method as 
\begin{equation}
\begin{aligned}
&\vm_t=\beta_{t}\vm_{t-1}+(1-\beta_{t}){\nabla}{f}_t(\dot{\vx}_t)\\
&\vx_{t+1}=\vx_t-\alpha_{t}\vm_{t}\\
&\dot{\vx}_{t+1}=\vx_{t}-(1+\eta_t)\alpha_{t}\vm_{t},
\end{aligned}
\label{RSGOri}
\end{equation}
where $\dot{\vx}_t$ is the observation point, $\alpha_t$, $\beta_{t}$, $\eta_{t}$ are configurable coefficients, and $\vm_0=0$. RSG is a generalization of NAG, because the observation distance $\eta_t$ can be configured to accommodate gradient noise, while it is set as $\beta_{t}$ in NAG to approximate the original NAG for exact gradients \cite{Sutskever2013}.

Both $\vx_t$ and $\dot{\vx}_t$ are required to update in (\ref{RSGOri}). To make the method more efficient, we simplify the update by approximation. Assume that the coefficients $\alpha_{t}$, $\beta_{1t}$, and $\eta_t$, change very slow between adjacent iterations. Substituting $\vx_{t}$ by $\dot{\vx}_t+\eta_{t-1}\alpha_{t-1}\vm_{t-1}$, we obtain the concise form of RSG, as
\begin{equation}
\begin{aligned}
&\vm_t=\beta_{t}\vm_{t-1}+(1-\beta_{t}){\nabla}{f}_t(\vx_t)\\
&\vx_{t+1}=\vx_t-\alpha_{t}\left((1-\mu_{t})\vm_{t}+\mu_{t}{\nabla}{f}_t(\vx_t)\right),\\
\end{aligned}
\label{RSG}
\end{equation}
where the observation factor $\mu_{t}=\eta_{t}(1-\beta_{t})/\beta_{t}$, and $\vx$ is used instead of $\dot{\vx}$ for simplicity. We further require $\beta_t,\mu_t\in[0,1)$.

In practical computation of RSG, the update form can be rearranged to a fast form as
\begin{equation}
\begin{aligned}
&\tilde{\vm}_{t}= \beta_{t} \tilde{\vm}_{t-1}+ {\nabla}{f}_t(\vx_t)\\
&\vx_{t+1} = \vx_t-\alpha_{t}\left(1-\beta_{t}\right)\left(1-\mu_{t}\right) \tilde{\vm}_{t} - \alpha_{t}\mu_{t}{\nabla}{f}_t(\vx_t),
\end{aligned}
\label{RSGReal}
\end{equation}
where only 3 scalar vector multiplications and 3 vector additions are required per iteration besides the gradient computation. The supplementary materials show that RSG is a more efficient equivalent form of ASGD \cite{Jain2018, kidambi2018}. Furthermore, it substantially reduces the cost for hyper-parameter tuning compared with ASGD by the analysis in Section 3, making it more feasible in real problems. 

Then, we construct the adaptive remote stochastic gradient (ARSG) by equipping RSG with a preconditioner modified from adaptive methods. The problem $\min_{\vx\in \mathcal{\hat{F}}}f(\vx)$ can be approximated locally as a stochastic quadratic optimization problem as
\begin{equation}
\min_{\vx\in \mathcal{\hat{F}}}\hat\Phi(\vx)=\frac{1}{2}{(\vx-\vx^*)^T} \mH (\vx-\vx^*),
\label{QuadOptProb}
\end{equation}
where $\mathcal{\hat{F}}$ is a local set of feasible points. Then, the approximate problem (\ref{QuadOptProb}) can be preconditioned as
\begin{equation}
\min_{\check{\vx}\in \mathcal{F}}\check\Phi(\check{\vx})=\frac{1}{2}{(\check{\vx}-\check{\vx}^*)^T}
\hat{\mV}_{t}^{-1/4} \mH\hat{\mV}_{t}^{-1/4}(\check{\vx}-\check{\vx}^*),
\label{PrecondProblem}
\end{equation}
where $\check{\vx}=\hat{\mV}_{t}^{1/4}\vx$, $\check{\vx}^*=\hat{\mV}_{t}^{1/4}\vx^*$, and $\hat{\mV}_{t}$ is a positive definite diagonal matrix. $\hat{\mV}_{t}^{-1/2}$ is called as the preconditioner. Update $\check{\vx_t}$ by $\nabla_{\check{\vx}}\check\Phi(\check{\vx_t})$ is equal to update $\vx$ by $\hat{\mV}_{t}^{-1/2}\nabla{\hat\Phi(\vx_t)}$.

Adaptive methods can be regarded as SGD or HB combined with adaptive preconditioners generated using the past gradients. By applying their preconditioners to RSG, we obtain generalize ARSG as
\begin{equation}
\begin{aligned}
&\vm_t=\beta_{t}\vm_{t-1}+(1-\beta_{t}){\nabla}{f}_t(\vx_t)\\
&{\vv}_t = {h_t}\left({\nabla}{f}_1(\vx_1),{\nabla}{f}_2(\vx_2),\cdots, {\nabla}{f}_t(\vx_t)\right) \\
&\vx_{t+1}= \Pi_{\mathcal{F},\mathrm{diag}({\vv}_t)} \left(\vx_t-\alpha_{t}\left((1-\mu_{t})\vm_{t}\right.\right.\\
&\left.\left.+\mu_{t}{\nabla}{f}_t(\vx_t)\right)/\sqrt{{\vv}_t}\right),\\
\end{aligned}
\label{General ARSG}
\end{equation}
where  $\vm_{0}=0$, the function $h_t$ can be defined following existing adaptive methods such as ADAM, AMSGRAD, RADAM and SADAM. In the following, we apply a nonincreasing preconditioner modified from AMSGRAD, that facilitates the convergence analysis. AMSGRAD \cite{Reddi2018} is given by 
\begin{equation}
\begin{aligned}
&\vm_t=\beta_{1t}\vm_{t-1}+(1-\beta_{1t}){\nabla}{f}_t(\vx_t)\\
&\vv_t=\beta_{2}\vv_{t-1}+(1-\beta_{2})({\nabla}{f}_t({\vx}_t))^2, \hat{\vv}_t=\max(\hat{\vv}_{t-1},\vv_t) \\
&\vx_{t+1}=\Pi_{\mathcal{F},\mathrm{diag}(\hat{\vv}_t)} \left(\left(\vx_t-\alpha_{t}\vm_{t}\right)/\left(\sqrt{\hat{\vv}_t}+\epsilon\right)\right),
\end{aligned}
\label{Precond AMSG}
\end{equation}
where $\beta_2$, $\epsilon\ll1$ are configurable coefficients, $\vv_0=\hat{\vv}_0=0$ \footnote{Although $\epsilon$ is not included in \cite{Reddi2018}, it cannot be omitted in implementation for stability.}. According to \cite{Tong2019} which shows that a relatively large $\epsilon$ is beneficial to promote the generalization by suppressing the rapid variation of the adaptive learning rate, we modified the preconditioner in (\ref{Precond AMSG}) by setting $\hat{\vv}$ with a relatively large initial value $\mathrm{diag}(\epsilon\mI)$. In order to save computation, the vector addition in the denominator is removed. 

Algorithm \ref{Algorithm1} shows the ARSG method \footnote{The hyper-parameters satisfy $\beta_{1t},\beta_2,\mu_t\in[0,1), 0<\epsilon\ll1$. As default $\beta_{1t}=0.999,\beta_2=0.99,\mu_t=0.1$. $\epsilon$ is $10^{-8}$ in the fast mode to accelerate convergence, and $10^{-3}$ in the fine mode to achieve better generalization. The updating is rearranged similar to the fast form (\ref{RSGReal}) in implementation to save computation.}. Compared with AMSGRAD, ARSG requires low computation overhead, as only one scalar vector multiplication per iteration, which is much cheaper than the gradient estimation. Besides, almost no more memory is required. In most cases, especially when weight decay is applied for regularization, which limits the norm of the parameter vectors, the projection can also be omitted in implementation to save computation. 

\begin{algorithm}[tb]
	\caption{ARSG Algorithm}
	\label{Algorithm1}
	\textbf{Input}: initial parameter $x_1$, coefficients $\{\alpha_{t}\}_{t=1}^{T}$,  $\{\beta_{1t}\}_{t=1}^{T}$, $\beta_2$,  $\{\mu_{t}\}_{t=1}^{T}$, $\epsilon$, iteration number $T$\\
	\textbf{Output}: parameter $\vx_t$
	\begin{algorithmic}[1] 
		\STATE Set $\vm_0=0$, $\vv_0=0$, and $\hat{\vv}_0=\mathrm{diag}(\epsilon\mI)$.
		\FOR {$t=1$ to $T-1$}
		\STATE $\vg_t={\nabla}{f}_t(\vx_t)$.
		\STATE $\vm_t=\beta_{1t}\vm_{t-1}+(1-\beta_{1t})\vg_t$.
		\STATE $\vv_t=\beta_{2}\vv_{t-1}+(1-\beta_{2})\vg_t^2$.
		\STATE $\hat{\vv}_t=\max(\hat{\vv}_{t-1},\vv_t), \hat{\mV}_t=\mathrm{diag}(\hat{\vv}_t)$. 
		\STATE $\vx_{t+1}=\Pi_{\mathcal{F},\sqrt{\hat{\mV}_t}}(\vx_t -\alpha_{t}((1-\mu_{t})\vm_{t}+\mu_{t}\vg_t)/\sqrt{\hat{\vv}_t})$.
		\ENDFOR
	\end{algorithmic}
\end{algorithm}

\section{Hyper-parameters selection guided by dynamic system analysis}

In RSG and ARSG, the configurable observation factor $\mu$ is costly to be selected by grid search. To reduce the huge costs, we investigate the convergence rate and error variance by dynamic system analysis, based on which the default value and an effective strategy to set the observation factor without grid search are proposed. In this section, we study the convergence of RSG (\ref{RSG}) in the local stochastic quadratic problem (\ref{QuadOptProb}), that can approximate both convex and non-convex problems. The hyper-parameters are set as constants, hence their time subscripts are ignored. 

Firstly, we model the optimization process as a dynamic system to reveal the convergence mechanism of RSG for different curvatures. Define $\vv_{e}$ as a unit eigenvector of the Hessian $\mH$, and the corresponding eigenvalue (curvature) is $\lambda$. We define the deviation
$s_t=\langle\vv_{e},\vx_t\rangle$, and the scalar momentum $\dot{v}_t=\langle\vv_{e},\vm_t\rangle$. According to (\ref{RSG}), the coefficients are updated as
\begin{equation}
\begin{aligned}
&\dot{v}_{t+1}={\beta}\dot{v}_{t}+(1-{\beta})\lambda(s_{t}+\delta_{t}),\\
&s_{t+1}=s_t-\alpha\beta(1-\mu){\dot{v}_t}
-\alpha\lambda(1-{\beta}(1-\mu))(s_{t}+\delta_{t}),
\label{CoeUpd}
\end{aligned}
\end{equation}
where the gradient error coefficient $\delta_{t}=\langle \zeta_t,\vv_e\rangle/\lambda$.

Substituting $\dot{v}_{t}$ by $v_{t}=\alpha\dot{v}_{t}$, and denote $\tau=\alpha\lambda$, we rewrite the update (\ref{CoeUpd}) as a dynamic system:
\begin{equation}
\begin{aligned}
&\begin{bmatrix}
v_{t+1}\\
s_{t+1}
\end{bmatrix}
=
A
\begin{bmatrix}
v_{t}\\
s_{t}
\end{bmatrix}
+
\tau\delta_{t}\vb,
\quad \text{where}\\
&\mA=
\begin{bmatrix}
\beta               &(1-{\beta})\tau\\
-\beta(1-\mu) &1-(1-{\beta}(1-\mu))\tau
\end{bmatrix}, \\
&\vb=\begin{bmatrix}
1-{\beta}\\
-(1-{\beta}(1-\mu))
\end{bmatrix}.
\label{DynSys}
\end{aligned}
\end{equation}
In (\ref{DynSys}) $[\vv_t,\vs_t]^T$ is called as the residual, and $\mA$ is called as the gain matrix, whose eigenvalues are
\begin{equation}
r_1, r_2=\frac{1}{2}\left(\rho \pm \sqrt{\rho^2-4\beta(1-\mu\tau)}\right),
\label{EigDynSys}
\end{equation}
where $\rho=1+\beta-\tau (1-\beta(1-\mu))$.

Denote the corresponding unit eigenvectors as $\vw_1$ and $\vw_2$, that are solved numerically since the expressions are too complicated. We also define the gain factor $r_g=\max(|r_1|,|r_2|)$, which is the convergence rate for specified $\tau$, $\beta$, and $\mu$ with exact gradients\footnote{Half a year after the configurable remote observation was proposed in our first preprint edition, Nakerst et al. \cite{Nakerst2020} also studied the gain factor for exact gradient independently. Their results are highly consistent with ours.}.

Define the coefficients $c_1,c_2,d_1,d_2$ satisfy
\begin{equation}
c_1\vw_1+c_2\vw_2=\vb,
d_1\vw_1+d_2\vw_2=
\begin{bmatrix}
v_1\\
s_1
\end{bmatrix}.
\label{CoeEigenV}
\end{equation}

From (\ref{DynSys}), (\ref{EigDynSys}) and (\ref{CoeEigenV}), we obtain
\begin{equation}
\begin{aligned}
&\begin{bmatrix}
v_{t+1}\\
s_{t+1}
\end{bmatrix}
=r_{1}^{t}d_1\vw_1+r_{2}^{t}d_2\vw_2 \\
&+\sum_{l=1}^{t+1}\left.\tau\delta_l\left(r_{1}^{t+1-l}c_1\vw_1+r_{2}^{t+1-l}c_2\vw_2\right)\right..
\label{DeductState}
\end{aligned}
\end{equation}

Assume that the noise satisfies $\E(\delta_i\delta_j)=0$ if $i\neq j$, $\delta_t=\sigma\delta$ where $\delta$ obeys the standard normal distribution and $\sigma$ is the standard deviation of $\delta_t$. From (\ref{DeductState}), we obtain the state expectation
\begin{equation}
\E(s_{t+1})=r_{1}^{t}d_1\vw_{1,2}+r_{2}^{t}d_2\vw_{2,2},
\label{StateEst}
\end{equation}
and the error variance limit
\begin{equation}
\begin{aligned}
&l_{\text{Var}}=\lim_{t\rightarrow+\infty}\Var(s_{t})/\delta^{2} \\
&=
\left(|c_{1}|^{2}|\vw_{1,2}|^{2}/(1-|r_{1}|^{2})+|c_{2}|^{2}|\vw_{2,2}|^{2}/(1-|r_{2}|^{2})\right.\\
&\left.+2 \operatorname{Re}\left(\overline{c_{1}} c_{2} \overline{\vw_{1,2}} \vw_{2,2}/(1-\overline{r_{1}} r_{2})\right)\right)\tau^{2},
\label{LimVarEst}
\end{aligned}
\end{equation}
if $\max(|r_1|,|r_2|)<1$.
 
Next, we study the convergence rate of RSG in quadratic problems with exact gradients. Analysis of eigenvalue (\ref{EigDynSys}) reveals the the following theorem:

\textbf{Theorem 1}. Consider the situation that RSG is used to solve problem (\ref{QuadOptProb}) with exact gradients. Assume that the Hessian $\mH$ is positive definite, and its conditional number $\kappa=\lambda_{\max}/\lambda_{\min}\gg1$, where $\lambda_{\max}$ and $\lambda_{\min}>0$ are maximal and minimal eigenvalues of $\mH$ respectively.  The hyper-parameters are set as follows: step size $\alpha=c_{\alpha}\sqrt{\kappa}/\lambda_{\max}$, momentum coefficient $\beta=1-c_{\beta}/\sqrt{\kappa}$, observation factor $\mu=c_{\mu}/\sqrt{\kappa}$, where $c_{\alpha}\le2/(c_{\beta}+c_{\mu})$, and $c_{\alpha},c_{\beta},c_{\mu}$ are $O(1)$ positive constants. Then, the convergence rate $r_c=\max_{\tau \in [\alpha\lambda_{\min},\alpha\lambda_{\max}]}(r_g)$ can be approximated as

\begin{equation}
r_c=\left\{
\begin{aligned}
&1-\left(c_{\beta}-\sqrt{c_{\beta}(c_{\beta}-4c_{\alpha})}\right)/(2\sqrt{\kappa}), \text{ if } 4 c_{\alpha} < c_{\beta}\\
&1-c_{\beta}/(2\sqrt{\kappa}), \text{ if } 4 c_{\alpha} \ge c_{\beta}.
\label{ConvRatQuad}
\end{aligned}
\right.
\end{equation}

The proof is given in the supplementary materials \footnote{Available at \url{https://github.com/rationalspark/NAMSG/blob/master/supplementary\%20materials.pdf}}. It should be noted that $r_c$ is the worst gain factor for all the eigenvalues of $\mH$, which is obtained when $\lambda=\lambda_{\min}$.  If $c_{\alpha} \ll c_{\beta}$, the convergence rate can be approximated as $r_c=1-c_{\alpha}/\sqrt{\kappa}$. The optimal hyper-parameters can be obtained by numerically solving  $\min_{\alpha,\beta,\mu}(r_g)_{\tau=\alpha\lambda_{\min}}, \text{ s.t. } (r_g)_{\tau=\alpha\lambda_{\min}}=(r_g)_{\tau=\alpha\lambda_{\max}}<1$. 

Theorem 1 shows that RSG can achieve a convergence rate of $1-O(1/\sqrt{\kappa})$ in quadratic problems with exact gradients. It is faster than the $1-O(1/\kappa)$ convergence rate of SGD without momentum (referred to as SGD0 hereinafter) in typical cases where $\kappa\gg1$. HB and NAG can be considered as special cases of RSG with $c_{\mu}=0$ and $c_{\mu}=c_{\beta}$, respectively. Consequently, they share the $1-O(1/\sqrt{\kappa})$ convergence rate of RSG.

Then, we study the combined effect of gain factor and gradient noise. For an eigenvalue $\lambda$, the residual $[\vv_t,\vs_t]^T$ can be decomposed along $\vw_1$ and $\vw_2$. Equation (\ref{DeductState}) shows that in each step the two components are multiplied by $r_1, r_2$ respectively, and the new gradient noise is introduced. In the early stage of training, when the initial error is dominant, the residual decays roughly exponentially, or oscillates while decaying exponentially since $r_1, r_2$ may be complex. As the training progresses, the accumulated gradient noise takes a greater proportion in the residual. Finally, the decaying of the residual is balanced by the new gradient error, leading to a saturation stage in which the loss flattens. The final loss expectation is
$l_{\text{Loss}}=\frac{1}{2}\lim_{t\rightarrow+\infty}\E(\lambda s_t^2)= \frac{1}{2}\lambda \delta^{2} l_{\text{Var}} 
\label{LimLoss}$. 

\begin{figure*}
	\begin{center}
		\includegraphics[width=1.0\textwidth]{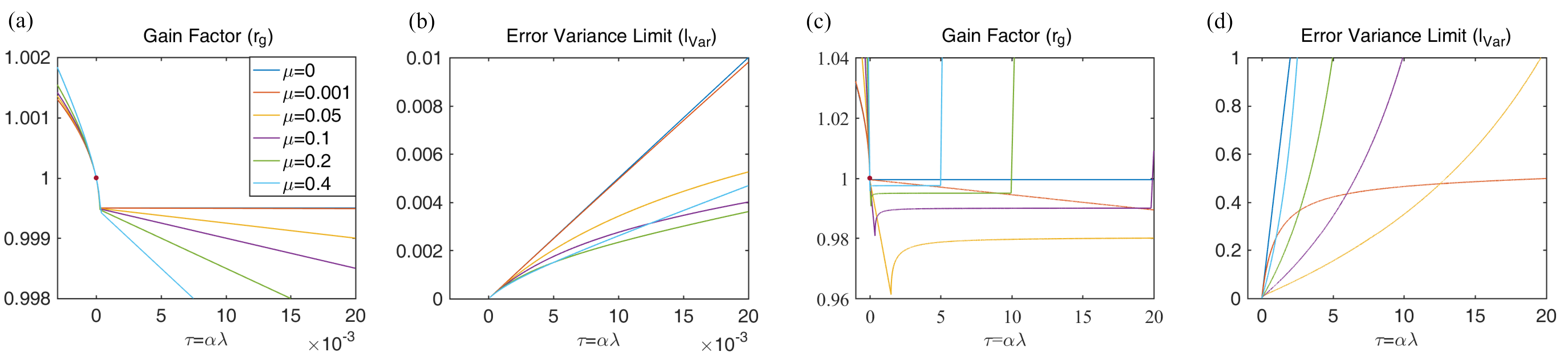}
	\end{center}
	\caption{The gain factor and the error variance limit of RSG when $\beta=0.999$: (a) and (b) for $\tau\in[-0.003,0.01]$, (c) and (d)  for $\tau\in[-1,20]$. The legend is shared.}
	\label{GainRSG}
\end{figure*}

Numerical simulation by the definition (\ref{LimVarEst}) for a large range of $\beta$ and $\mu$ (e.g. Figure \ref{GainRSG}) shows that $l_{\text{Var}}$ increases with $\tau$. According to Theorem 1, when $4 c_{\alpha} \ge c_{\beta}$, the convergence rate $r_c$ is determined by $c_{\beta}$, and a small $\alpha$ is preferred to suppress the noise. When $4c_{\alpha} < c_{\beta}$, $r_c$ is a decreasing function of $c_{\alpha}$, there would be a compromise between a large $\alpha$ to bring down $r_c$ and a small $\alpha$ to suppress the noise. Comparison between RSG and SGD0 further reveals the effect of noise. The maximum step size of SGD0 is $O(1/\lambda_{\max})$. However, Theorem 1 demonstrates that RSG requires the step size $\alpha=O(\sqrt{\kappa}/\lambda_{\max})$ to obtain the $1-O(1/\sqrt{\kappa})$ convergence rate, which generates an error variance expectation much larger than SGD0, leading to unacceptable error when $\kappa\gg1$. Consequently, the noise prevents RSG (including HB and NAG) from achieving the $1-O(1/\sqrt{\kappa})$ convergence rate when the gradient is noisy, and a practical step size should be much less than $O(\sqrt{\kappa}/\lambda_{\max})$.

Theorem 1 also shows that the observation factor $\mu$ has little impact on $r_c$ directly when $\kappa\gg1$. Instead, a relatively large $\mu$ limit the largest step size in the assumption, and may enlarge $r_c$ with exact gradient when $4c_{\alpha} < c_{\beta}$. But it does not harm the convergence in most real problems, where the practical step size limited by noise is much smaller than the $O(\sqrt{\kappa}/\lambda_{\max})$ bound in the assumption.

Finally, we study the observation factor strategy to decrease both the gain factor $r_g$ and the error variance limit $l_{\text{Var}}$ for different curvatures. Since the convergence rate $r_c$ only represents the worst case, we further study the convergence properties for different eigenvalues by Figure \ref{GainRSG}, which presents $r_g$ and $l_{\text{Var}}$ for different $\tau=\alpha\lambda$ and $\mu$ \footnote{To take the advantage of fast convergence for exact gradients, the momentum factor $\beta$ should be close to 1 according to $\beta=1-O(1/\sqrt{\kappa})$ in the assumption of Theorem 1. So we set the default value of $\beta$ to $0.999$.}. It is shown that a large $\beta$ and a proper $\mu$ ensure a large convergence domain ($0<\tau<2$ is required for convergence in SGD0). When $\alpha$ is specified, the eigenvalue $\lambda$ is proportional to $\tau$. For a very small positive $\lambda$, the observation factor $\mu$ has little effect on $r_g$, that is consistent with the theoretical analysis. When $\lambda$ is slightly larger, a relatively large $\mu$ (e.g. $0.1$) results in significantly improvement in both $r_g$ and $l_{\text{Var}}$, while HB ($\mu=0$) and NAG ($\mu=0.001$) still converge very slow for these eigenvalues, impeding the entire training process. In non-convex problems when $\lambda<0$, a relatively large $\mu$ also accelerates the divergence. A small $\mu$, such as $\mu=0.001$ (NAG), generates large $l_{\text{Var}}$ for small $\lambda$ (Figure \ref{GainRSG} (b)), that limits its improvement upon HB. It does improves $r_g$ and $l_{\text{Var}}$ for very large $\tau$, but the convergence for large $\lambda$ is not the bottleneck and a too large step size $\alpha$ is not applicable since it amplifies $l_{\text{Var}}$ for all the eigenvalues. Too large $\mu$ (e.g. 0.4) is not applicable because it generates large $l_{\text{Var}}$ for a very large range of $\lambda$ (Figure \ref{GainRSG} (b), (d)). We also studied $\beta=0.9,0.99$, and observed that $\mu\in[0.05,0.2]$ is a feasible range to decrease both $r_g$ and $l_{\text{Var}}$. Consequently, we set $\mu=0.1$ as the default value. Only the step size $\alpha$ is left for grid search in RSG.

Figure \ref{GainRSG} shows for a tiny range of $\tau>0$, the gain factor $r_g$ decreases with $\tau$ rapidly and is almost independent to $\mu$. Then $r_g$ decreases with $\tau$ roughly linearly to a minimum which is improved when $\mu$ is slightly small. Consequently, in the early stage of training when the initial error is dominant, a large step size $\alpha$ and a small $\mu$ can reduce $r_g$ of tiny positive eigenvalue $\lambda$, and they also accelerate the divergence of tiny negative $\lambda$ in non-convex settings. However, the large $\alpha$ enlargers $l_{\text{Var}}$. When the loss flattens, decaying $\alpha$ is required to reduce noise, and increasing $\mu$ can lower $r_g$ for relatively small $\lambda$. By the concept we propose a hyper-parameter policy named observation boost (OBSB). The policy performs grid search for a small portion of iterations using a small $\mu$ to select an optimal initial $\alpha$. When the loss flattens, it doubles $\mu$, and scales $\alpha$ proportional to $\text{argmin}_{\tau}r_g$. After the adjustment of $\mu$ and $\alpha$ (called boosting), $\lambda$ to minimize $r_g$ is unchanged. OBSB is different from vanilla learning rate decay. In the grid search with $\beta$ close to $1$ and a small $\mu$, the step size $\alpha$ obtained is generally large. After boosting $\alpha$ is still relatively large, so that it suffers less from premature caused by too small step size. The default initial $\mu$ is 0.05, and $\alpha$ is roughly divided by 4 in the boosting.

Because ARSG can be regarded as a preconditioned version of RSG and the preconditioner varies slow when $t$ is large, the observation factor strategy of RSG is directly applied to ARSG for simplicity. 

\section{Convergence Analysis}
In this section, we provide a $O(1/\sqrt{T})$ convergence bound of ARSG in non-convex optimization, and further present a $O(\log(T))$ regret bound for strongly convex optimization.

We make the following assumptions \cite{Chen2018on}.

\textbf{A1}: $f$ is differentiable and has $L$-Lipschitz gradient, i.e. $\forall \vx,\vy$,
$\|\nabla f(\vx) - \nabla f(\vy) \| \leq L \|\vx-\vy\|.$ It is also lower bounded, i.e. $f(\vx) > - \infty$.

\textbf{A2}: The feasible set $\mathcal{F}$ has bounded diameter $D_{\mathcal{F}}$, the gradients and noisy gradients are also bounded, i.e. $\|\vx-\vy\|\le{D_{\mathcal{F}}}$, $\|\nabla f(\vx)\|_{1}, \|\nabla f_{t}(\vx)\|_{1}{\le}G_{1}$, $\|\nabla f(\vx)\|, \|\nabla f_{t}(\vx)\|{\le}G_{2}$, for any $\vx,\vy\in{\mathcal{F}}$. 

\textbf{A3}: Gradient noise is zero-mean and independent. i.e. $\E[\zeta_t] = 0$ and $\zeta_i,\zeta_j$ are independent if $i\neq j$.

Firstly, we present the convergence bound of ARSG in non-convex settings by the following theorem.

\textbf{Theorem 2.} 
For iteration budget $T$, suppose that the hyper-parameters are $\alpha_t = \dot{\alpha}/\sqrt{T}>0$, $0 \le \beta_{1,t}=\beta_{1} < 1$, $0 \le \mu_{t} =\mu <1$, for $t\in[T]$. ARSG (Algorithm \ref{Algorithm1}, where the projection operation is omitted) yields  
\begin{equation}
\begin{aligned}
\label{BndARSG}
&\min_{t\in[T]} E\left[ \|\nabla f(x_t)\|^2\right] 
\leq  
D_1 \frac{\dot{\alpha}}{\sqrt{T}\epsilon}
+ D_2 \frac{1}{\dot{\alpha}\sqrt{T}}\\
&+ (D_3 d +D_4) \frac{1}{T \sqrt{\epsilon}}
+ (D_5 d+D_6) \frac{\dot{\alpha}}{T^{3/2} \epsilon },
\end{aligned}
\end{equation}
where the coefficients $D_1,D_2,D_3,D_4,D_5,D_6$ are constants independent of $T,\epsilon, \text{ and }d$.

Theorem 2 shows that in non-convex settings ARSG yields $O(1/\sqrt{T})$ convergence rate with constant hyper-parameters, while the terms related with the dimension $d$ enjoys $O(1/T)$ convergence rate. In the bound (\ref{BndARSG}), the term $D_2/(\dot{\alpha}\sqrt{T})$ represents the influence of initial error and momentum, that is independent of $\epsilon$. The other terms are related to the error introduce by the adaptive preconditioner, therefore they are sensitive to $\epsilon$. In the supplementary materials, we show that AMSGRAD shares the same form of bound under the assumption of Theorem 2, except for $\mu=0$, $\epsilon$ is substituted by $\epsilon^2$ and the coefficients are slightly different. Compared with AMSGRAD, ARSG improves the bound by modifying the definition of $\epsilon$. Owning to the remote gradient observation, it also improves the coefficients in typical cases where $d \gg 1, 1-\beta_{1} \ll 1, L \gg 1, G_{2}\gg 1, \epsilon \ll 1$.

Then, we demonstrate that the convergence rate of ARSG can be substantially improved in strongly convex settings. We evaluate the bound by regret, defined as $R_T=\sum_{t=1}^{T} (f_t(\vx_t)-f_t(\vx^*))$. Theorem 3 indicates ARSG can further achieve $O(\log(T))$ regret bound with $O(1/t)$ step size \cite{Bottou2018,Wang2019}. 

\textbf{Theorem 3}. Assume that $\forall x_1,x_2\in{\mathcal{F}}$,
$f_t\left(\vx_{1}\right)$ $ \geq f_t\left(\vx_{2}\right)+\nabla f_t\left(\vx_{2}\right)^{\top}\left(\vx_{1}-\vx_{2}\right)
+\frac{\lambda}{2}\left\|\vx_{1}-\vx_{2}\right\|^{2}$, where $\lambda$ is a positive constant.
Let $\{\vx_t\}$, $\{\vv_t\}$ and $\{\hat{\vv_t}\}$ be the sequences obtained from Algorithm \ref{Algorithm1}. The initial step size $\alpha \geq \max_{i\in\{1,\cdots,d\}}$ $\left(t \hat{\vv}_{t,i}^{1 / 2}-(t-1) \hat{\vv}_{t-1,i}^{1 / 2}\right) /\left(\left(1-\beta_{1}(1-\mu)\right) \lambda\right)$, $\alpha_t=\alpha/{t}$, $0\leq\beta_{t}=\beta_1/t^2<1$, $\gamma=\beta_1/\sqrt{\beta_2}<1$, $1-\beta_{1} \leq \mu_{t}=\mu <1$, $0 < \epsilon \ll 1$, for $t\in[T]$, and $\vx\in{\mathcal{F}}$. We have the following bound on the regret
\begin{equation}
\begin{aligned}
&R_{T} \leq \left(\frac{\alpha G_{1}}{\sqrt{1-\beta_{2}}}\left(\frac{3}{2} \frac{\beta_{1}^{2}}{\left(1-\beta_{1}\right)(1-\gamma)}+\mu^{2}\right)\right. \\
&\left.+\frac{(1-\mu) \beta_{1} D_{\mathcal{F}}^{2}}{2 \alpha \sqrt{\epsilon}} \right)\frac{1+\log (T)}{1-\beta_{1}(1-\mu)}.
\label{RegretARSGStronglyConvex}
\end{aligned}
\end{equation}
The bound (\ref{RegretARSGStronglyConvex}) shows that in strongly convex settings ARSG can achieve $O(\log(T)/T)$ convergence rate, which is independent of the dimension $d$. Under the same assumption, AMSGRAD shares the bound (\ref{RegretARSGStronglyConvex}) except for $\mu=0$, and $\epsilon$ is substituted by $\epsilon^2$. ARSG improves the bound by redefining $\epsilon$, and also improves the coefficients in typical cases where $1-\beta_{1} \ll 1, \epsilon \ll 1$.

The proofs are given in the supplementary materials. It should be noted that the bounds in Theorem 2 and 3 only represent the worst cases, while the acceleration of ARSG mainly comes from utilizing second order information in the local stochastic quadratic approximation as shown in Section 3.


\section{Experiments}

\begin{figure*}
	\centering
	\includegraphics[width=1.0\textwidth]{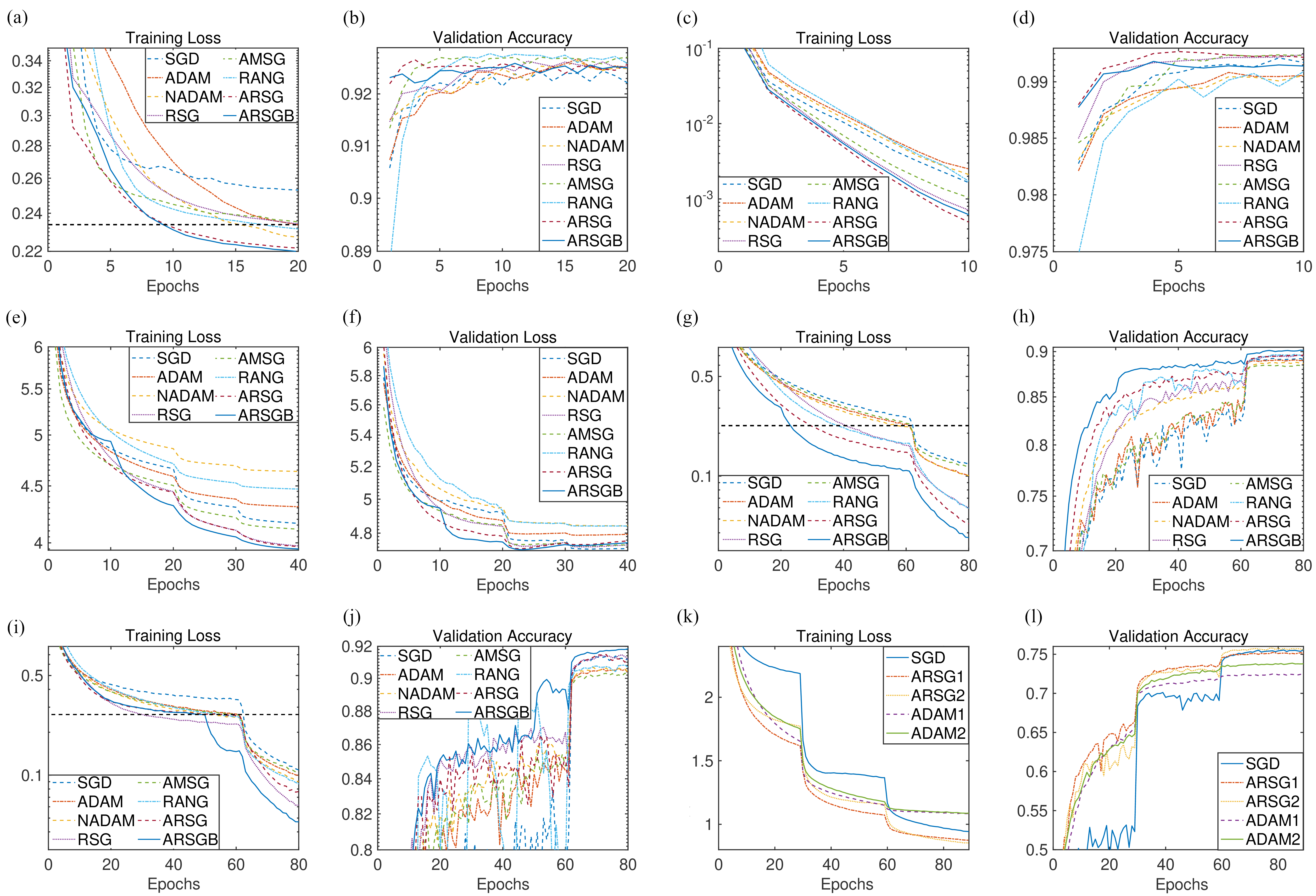}
	\caption{Results of experiments: (a), (b) logistic regression on MNIST; (c), (d) CNN on MNIST; (e), (f) LSTM on WikiText-2; (g), (h) ResNet-20 on CIFAR-10 (fast mode); (i), (j) ResNet-20 on CIFAR-10 (fine mode); (k), (l) ResNet-50 on ImageNet.}
	\label{Experiments}
\end{figure*}

In this section, we present experiments to evaluate the performance of RSG, ARSG and ARSG with the OBSB policy (denoted as ARSGB). The experiments are performed on MNIST \cite{Lecun1998}, WikiText-2 \cite{Merity2016}, CIFAR10 \cite{Krizhevsky2009}, and ImageNet \cite{Deng2009} datasets \footnote{The experiments on MNIST and CIFAR10 are carried out with MXNET \cite{Chen2015}, the others with PyTorch \cite{Pytorch2019}.}. On the first 3 datasets, the results are compared  with SGD with momentum (HB) \cite{Polyak1964}, and popular adaptive methods, such as ADAM \cite{Kingma2015}, NADAM \cite{Dozat2016}, AMSGRAD \cite{Reddi2018}, and RANGER \cite{Wright2019} \footnote{To avoid overlapping, AMSGRAD and RANGER are abbreviate to AMSG and RANG in the figures.}. For RSG, ARSG and ARSGB, only the step size is selected by grid search, and other hyper-parameters are set to their default values, while exhaustive grid search are applied to optimize the performance of the compared algorithms. On ImageNet, we compared the performance of SGD, ARSG, and ADAM. Details of the experiments, such as the description of datasets, the model structures, the preprocessing, and the hyper-parameters selection, are presented in the supplementary materials. The experiment results are shown in Figure \ref{Experiments}, which are the mean of $5$ runs (except for the experiments on ImageNet, they run only once).

The experiments on MNIST (Figure \ref{Experiments}(a-d)) and WikiText-2 (Figure \ref{Experiments}(e, f)) mainly compare the convergence speed. The results show that ARSG and ARSGB converge much faster than former methods. For training logistic regression on MNIST (Figure \ref{Experiments}(a)), which is a typical example of convex optimization, ARSG and ARSGB are more than $50$\% faster than ADAM. RSG is also much faster than SGD. It further outperforms popular adaptive method in many cases, even if it is not equipped with an adaptive preconditioner. Besides, RSG, ARSG and ARSGB consume much less trials in the grid search for hyper-parameters selection since they merely search for the step size. 

The experiments on CIFAR-10 (Figure \ref{Experiments}(g-j)) compare both the fastest convergence and the best generalization. In the fast mode to minimizes the training loss, ARSG and ARSGB are much faster than other methods. RSG is as fast as RANGER, and faster than former methods. In the fine model to maximizes generalization, larger step sizes are selected in grid search, showing that relatively large steps are beneficial to generalization by extending exploration at the cost of slower convergence. ARSGB achieves the best generalization. RSG and ARSG generalizes as good as SGD, and much better than former adaptive methods. The max validation accuracies of SGD, ADAM, NADAM, AMSGRAD, RANGER, RSG, ARSG, and ARSGB are $0.9151\pm0.0021, 0.9079\pm0.0026, 0.9080\pm0.0022, 0.9045\pm0.0013, 0.9102\pm0.0022, 0.9161\pm0.0018, 0.9157\pm0.0019, 0.9196\pm0.0014$, where $\pm$ denotes the standard deviation. ARSG and ARSGB also converges faster than other methods.  

The experiments on ImageNet (Figure \ref{Experiments}(k, l)) show that ARSG converges fast and generalizes well at the same time. ADAM1 ($\alpha=0.0001$) and ADAM2 ($\alpha=0.0002$) converge faster than SGD, but generate large generalization gap. ARSG2 ($\alpha=0.02$) is faster than ADAM1 and ADAM2 in most of the epochs, and surpasses SGD in generalization. With a smaller step size that jeopardizes the generalization, ARSG1 ($\alpha=0.01$) still generalizes close to SGD, while  it converges much faster than ADAM1 and ADAM2. The max top 1 validation accuracies of SGD, ADAM1, ADAM2, ARSG1 and ARSG2 are $0.7559, 0.7249, 0.7386, 0.7524, 0.7579$. 

\section{Conclusions}

We present the RSG method, which computes the gradients at configurable remote observation points. It is further combined with elementwise preconditioner to construct ARSG for acceleration. Convergence analysis shows that in non-convex settings ARSG yields $O(1/\sqrt{T})$ convergence rate, which can be improved to $O(\log(T)/T)$ in strongly convex settings. Numerical experiments demonstrate that ARSG outperforms popular adaptive methods, such as ADAM, NADAM, AMSGRAD, and RANGER, in both convergence speed and generalization \footnote{In further studies, we find RSG coincides with QHM \cite{Ma2018}. They show the excellence of QHM (RSG) by extensive grid search.  However, the main difficulty in application of QHM is hyper-parameter selection, where 3 parameters require tuning. Our theoretical contributions focus on the dynamic system analysis to reveal why RSG converges faster than SGD and NAG, and guide the hyper-parameter selection. The experiments show that RSG and ARSG with the default hyper-parameters perform closely to QHM with optimal hyper-parameters obtained through extensive grid search, and they are much faster than SGD and NAG. The default hyper-parameters of RSG also expedites convergence compared with those of QHM (obtained from experiments) in our experiments. Besides, the key ideas of the dynamic system analysis are similar to \cite{Gitman2019}, but they are contemporary works and our former reprint edition appears earlier.}

{\small
\bibliographystyle{ieee_fullname}
\bibliography{arsgbib}
}

\end{document}